\documentclass[conference]{IEEEtran}
\IEEEoverridecommandlockouts
\usepackage{cite}
\usepackage{amsmath,amssymb,amsfonts}
\usepackage{algorithmic}
\usepackage{graphicx}
\usepackage{textcomp}
\usepackage{xcolor}

\usepackage{booktabs}       
\usepackage{amsfonts}       
\usepackage{nicefrac}       
\usepackage{microtype}      
\usepackage{lipsum}
\usepackage[ruled,vlined,linesnumbered]{algorithm2e}
\usepackage{setspace}
\usepackage{amsmath}
\usepackage{amsthm}

\usepackage{graphicx}
\usepackage{subcaption}
\usepackage{xcolor}
\usepackage{cite}
\newcommand{\matr}[1]{\mathbf{#1}} 

\newcommand{\fedave}{\textsf{FedAvg}}
\newcommand{\fedavec}{\textsf{FedAvg (Contextual)}}
\newcommand{\fedprox}{\textsf{FedProx}}
\newcommand{\fedproxc}{\textsf{FedProx (Contextual)}}
\newcommand{\folb}{\textsf{FOLB}}
\newcommand{\mnist}{\emph{MNIST}}
\newcommand{\emnist}{\emph{FEMNIST}}
\newcommand{\syniid}{\emph{Synthetic\_IID}}
\newcommand{\synone}{\emph{Synthetic\_1\_1}}

\newcommand{\norm}[1]{\left\lVert#1\right\rVert_2}
\newcommand{\fl}{\textsf{FL}}
\newcommand{\w}{\mathbf{w}}
\newcommand{\e}{\mathbf{e}}
\newcommand{\E}{\mathbb{E}}
\newcommand{\x}{\mathbf{x}}

\allowdisplaybreaks

\newtheorem{theorem}{Theorem}
\newtheorem{definition}{Definition}

\newtheorem{property}{Property}
\newtheorem{proposition}{Proposition}

\def\BibTeX{{\rm B\kern-.05em{\sc i\kern-.025em b}\kern-.08em
    T\kern-.1667em\lower.7ex\hbox{E}\kern-.125emX}}
\begin{document}

\title{Contextual Model Aggregation for Fast and Robust Federated Learning in Edge Computing
}

\author{%
	\IEEEauthorblockN{Hung T. Nguyen\IEEEauthorrefmark{1},
		H. Vincent Poor\IEEEauthorrefmark{1},
	and Mung Chiang\IEEEauthorrefmark{3},}
	\IEEEauthorblockA{\IEEEauthorrefmark{1}%
		Princeton University,
		\{hn4,poor\}@princeton.edu}
	\IEEEauthorblockA{\IEEEauthorrefmark{3}%
		Purdue University,
		chiang@purdue.edu}
}

\maketitle

\begin{abstract}
Federated learning is a prime candidate for distributed machine learning at the network edge due to the low communication complexity and privacy protection among other attractive properties. However, existing algorithms face issues with slow convergence and/or robustness of performance due to the considerable heterogeneity of data distribution, computation and communication capability at the edge.
In this work, we tackle both of these issues by focusing on the key component of model aggregation in federated learning systems and studying optimal algorithms to perform this task. Particularly, we propose a contextual aggregation scheme that achieves the optimal context-dependent bound on loss reduction in each round of optimization. The aforementioned context-dependent bound is derived from the particular participating devices in that round and an assumption on smoothness of the overall loss function. 
We show that this aggregation leads to a definite reduction of loss function at every round.
Furthermore, we can integrate our aggregation with many existing algorithms to obtain the contextual versions.
Our experimental results demonstrate significant improvements in convergence speed and robustness of the contextual versions compared to the original algorithms. We also consider different variants of the contextual aggregation and show robust performance even in the most extreme settings.
\end{abstract}

\begin{IEEEkeywords}
Federated learning, distributed optimization, robust learning, edge computing systems
\end{IEEEkeywords}

\section{Introduction}
Federated learning (\fl) was first introduced in \cite{mcmahan17} as a distributed machine learning algorithm and has been largely accepted as a framework of choice in edge learning systems due to the numerous nice properties including low communication complexity and data privacy protection, which has been in the spotlight in recent years. This coupled with the explosive increases of smart devices, e.g., smartphones, wearables, autonomous vehicles, in both number and capability, has fueled substantial research efforts \cite{li19survey,yang2019federated,kairouz2019advances} to understand and develop \fl{} algorithms with better convergence properties and higher adaptivity to the inherent heterogeneity in edge learning systems. It is also on the same line with the recent trend of migrating from cloud to edge for better latency and improving user experience.

\begin{figure}[t!]
	\includegraphics[width=\linewidth]{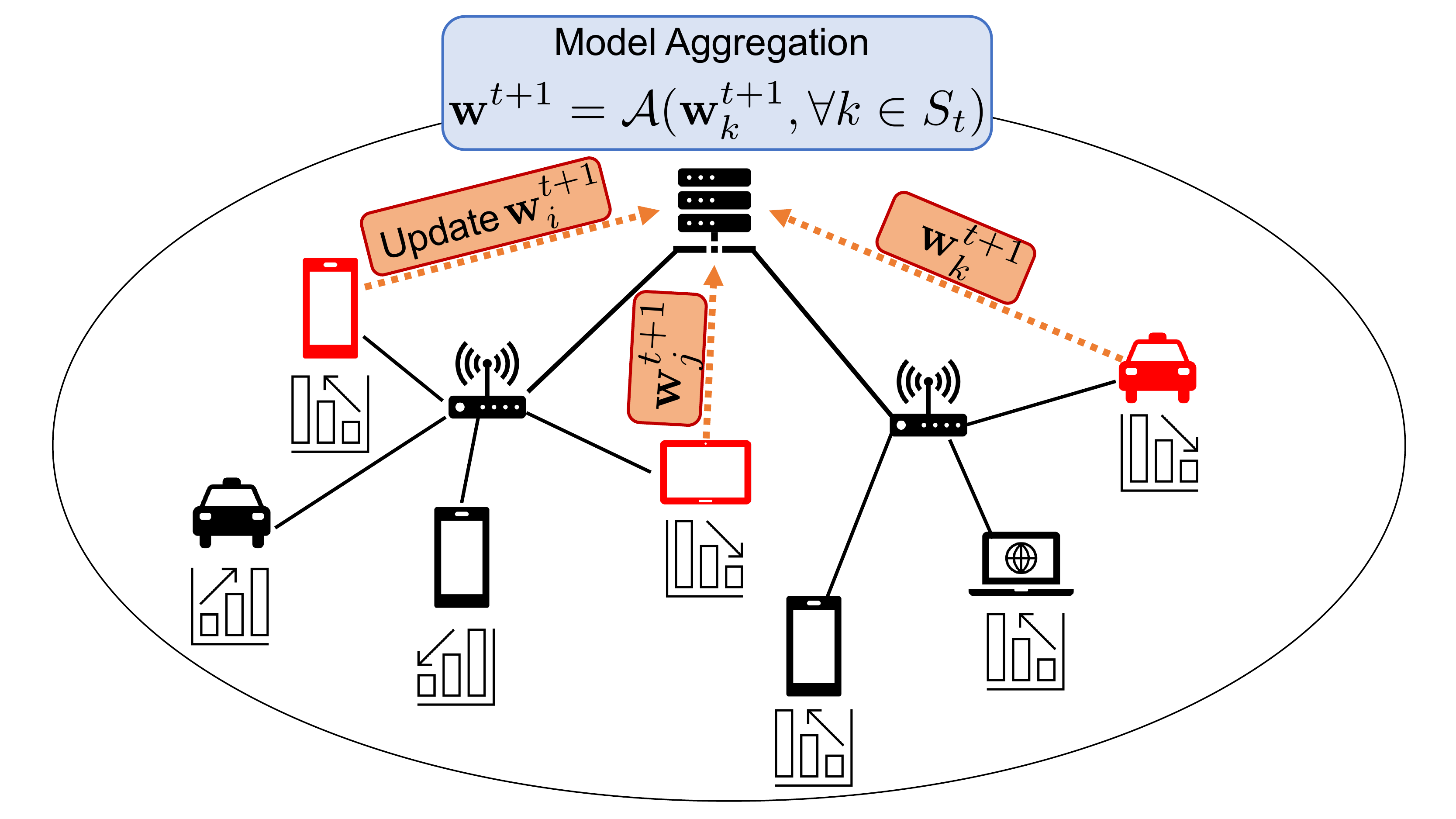}
	\caption{Federated learning framework with the key aggregation scheme $\mathcal{A}$ to combine received parameters updates from a small set of devices selected in round $t$ of optimization. Our focus is to investigate optimal aggregation that quickly reduces the overall loss and provides robust performance in the presence of statistical and system heterogeneity.}
	\label{fig:fl_model}
	\vspace{-0.15in}
\end{figure}

The \fl{} framework assumes a network of $N$ connected heterogeneous devices, each of which owns a separate set of private local data and wishes to learn a machine learning model collaboratively with other devices without sharing any data. In addition, there is a central server that can communicate, aggregate and exchange model parameters with all the devices. A typical \fl{} algorithm unfolds in multiple rounds, and in each round $t$, a small set $S_t$ of $K$ random devices are selected to perform local optimization from the current parameters and send updates to the central server for aggregation. The newly aggregated parameters form the latest model and will be updated upon in the next round. Figure~\ref{fig:fl_model} illustrates the components of \fl{} framework and the focus of our work. We can identify multiple inherent sources of heterogeneity in this framework: 1) (statistical) non-i.i.d. data across devices; 2) (system) different computing resources at devices; and 3) (system) varying communication latency between central server and devices. Thus, on top of challenges in traditional distributed optimization in the cloud, we need to address issues associated with the inherent heterogeneity in \fl{} setting.

\fedave{}\cite{mcmahan17} was the first \fl{} algorithm and has demonstrated outstanding performance in different tasks,
however, its performance degrades greatly in heterogeneous environments, and much development has been made to analyze and improve the convergence rate of \fedave{} \cite{li2019convergence,yu2019parallel,khaled2020tighter,tian2019,dinh2020personalized,karimireddy2020scaffold,nguyen2020fast}. Most of these works focus on modifying local optimizers to mitigate data heterogeneity and prove promising convergence rates under assumptions on smoothness, bounded dissimilarity between data distribution across devices, and/or function convexity. Most notably, \fedprox{} \cite{tian2019} incorporates a proximal term in each local loss function to penalize drastic changes of parameters in each round and, therefore, prevent majorly conflicting updates from devices with totally different data. Very recently, \folb{} \cite{nguyen2020fast} focuses on the aggregation scheme at the central server and integrates a weighted averaging by inner product between local and global gradients. However, as shown in \cite{tian2019,nguyen2020fast}, although recent methods provide better performance than \fedave{}, they struggle with keeping the results robust throughout the optimization process and experience wide fluctuations, even in consecutive rounds.

In this work, we aim at addressing both convergence and robustness issues in federated learning by studying optimal aggregation algorithms at the central server. Given the inherent heterogeneity at multiple scales, the aggregation step is the key to distilling local updates from selected devices and forming a robust model parameters. We consider a general form of model aggregation and derive a context-dependent bound based on the particular set of selected devices, referred to as \emph{context} of that round hereafter, and a minimal assumption of the overall loss function being smooth (common in both convex and non-convex optimizations). Then, the optimal setting of model aggregation is derived by optimizing the context-dependent bound and is used to form the next global model parameters.
We prove a definite reduction off the loss function with the magnitude depending on the local updates
and aggregation variables derived by our technique. We also study the bound in expectation over all random choices of participants and derive a corresponding aggregation.

In the experiments, we consider multiple variants of our aggregation, which differ in hyper-parameter settings, and show 
consistently high performance and robustness between variants. We also integrate the contextual aggregation scheme with existing \fl{} algorithms and reveal significant improvements in terms of convergence speed and much better robustness property. The contextual versions can reduce the number of optimization rounds to achieve the same accuracy levels effectively by a factor of three or more across various settings and datasets. We summarize our contributions as follows:
\begin{itemize}
	\item We study a general form of model aggregation at the central server in federated learning and propose an optimal aggregation in a general form to optimize a context-dependent bound based on the specific participating devices in each round. The context-dependent bound only relies on the particular set of devices and a minimal smoothness assumption of the overall loss function.
	\item We show that the proposed contextual aggregation results in definite loss reduction with the magnitude depending on local updates. We extend our scheme to the case of optimizing the expected loss reduction over random device selection.
	\item We examine variants of the contextual aggregation and perform extensive experiments in various settings and datasets to demonstrate faster convergence and much higher robustness when integrating the new scheme into existing federated learning algorithms.
\end{itemize}

\section{Preliminaries}
We introduce the system model (Subsection~\ref{subsec:model}), different types of heterogeneity (Subsection~\ref{subsec:hetero}), and details of important federated learning algorithms (Subsection~\ref{subsec:fed_alg}) that we consider most related and compare with throughout this work.
\subsection{System models and problem definition}
\label{subsec:model}
We consider a network of $N$ devices, indexed from 1 to $N$. Each device $k$ contains a private dataset $\mathcal{D}_k$ and each data point $d \in \mathcal{D}_k$ is a pair of feature vector $\x_d$ and label $y_d$ (in classification or regression task). Additionally, there is a central server that is connected with all $N$ devices. The problem of interest is to learn a common machine learning model collaboratively using all the available data in the entire network without sharing any (private) data. In other words, the common model works well on all the devices and achieves better generalization than one trained on a single local dataset. We also wish to incur minimal communication between devices and server to avoid interfering with network operations.


More specifically, the machine learning model is identified by a parameter vector $\w$ in $n$-dimensional space and a loss function $f(\w) = (1 / |\mathcal{D}|) \sum_{d \in \mathcal{D}} l(\w, x_d, y_d)$ to be minimized, where $\mathcal{D}$ is the combined dataset available $\mathcal{D} = \cup_{k = 1}^{N}\mathcal{D}_k$, and $l(\w, \x_d, y_d)$ represents the particular error between $y_d$ and the model output on $\x_d$(e.g., the squared distance). Thus, we aim to find $\w$ that minimizes $f(\w)$ over the data $\mathcal{D}$  in the network. In federated learning setting, this minimization is not performed directly, as each device $k$ only has access to $\mathcal{D}_k$. We further define $F_k(\w) = (1 / |\mathcal{D}_k|) \sum_{d \in \mathcal{D}_k} l(\w, \x_d, y_d)$ as the local loss function at $k$ over $\mathcal{D}_k$. If we assume that $|\mathcal{D}_i| = |\mathcal{D}_j| \; \forall i,j$, i.e., each device contains the same number of data points, we can simplify and express the optimization as an average over $F_k(\w)$:
\begin{equation}
	\label{eq:decomp}
	\min_\w f(\w) = \min_\w \frac{1}{N}\sum_{k = 1}^N F_k(\w).
\end{equation}
In general cases, devices may possess different amounts of data, e.g., due to data heterogeneity and varying compute capabilities. In such cases, we can replace the factor $1 / N$ with $p_k = |\mathcal{D}_k| / |\mathcal{D}|$ for a weighted average of the $F_k(\w)$ \cite{tian2019,tu20}.

Federated learning algorithms differ in how \eqref{eq:decomp} is solved. In our case, we will assume that a central server is available to orchestrate the learning across the devices. Such a scenario is increasingly common in fog or edge computing systems, where an edge server may be connected to edge devices, e.g., in a smart factory \cite{tu20}. We will next introduce the standard algorithms for federated learning in these environments.

\subsection{Inherent heterogeneity in edge learning systems}
\label{subsec:hetero}
We consider the presence of various heterogeneity sources commonly manifested in practical edge computing systems:
\begin{itemize}
	\item[1.] \emph{Data heterogeneity}: The generating distributions of data in devices are different. Data in a particular device depend on the distinct factors, e.g., usage, geographic location.
	Prior works usually need to assume some form of bounded dissimilarity across devices.
	\item[2.] \emph{Computational heterogeneity:} The computing resources at different devices vary that leads to the fact that some devices can perform local optimization to near optimality while others only change the parameters slightly or may not optimize at all.
	\item[3.] \emph{Communication heterogeneity:} The latency between devices and the aggregation server is also non-homogeneous.
\end{itemize}

\subsection{Federated learning algorithms and context definition}
\label{subsec:fed_alg}

\begin{algorithm}[h]
	\caption{\fedave{} - Federated learning with simple averaging.}
	\label{alg:fed_avg}

	\DontPrintSemicolon
	
	\setstretch{1.2}
	\SetKwInOut{Input}{Input}\SetKwInOut{Output}{output}
	\Input{$K, T,  \w^0, N$}
	
	\For{$t = 0, \dots, T-1$}{
		Server samples a small set $S_t$ of $K$ devices uniformly at random. \\
		Server sends $\w^t$ to all devices $k \in S_t$. \\
		Each device $k \in S_t$ finds a $\w_k^{t+1}$ that optimizes the local loss function at device $k$. \\
		Each device $k \in S_t$ sends $\w_k^{t+1}$ back to the server.\\
		Server aggregates the $\w_k^{t+1}, \forall k \in S_t$ according to \eqref{eq:sim_avg} to form a $\w^{t+1}$.
	}
\end{algorithm}

The first federated learning algorithm, \fedave{} \cite{mcmahan17}, is summarized in Algorithm~\ref{alg:fed_avg}. Essentially, \fedave{} selects random $K$ devices in each round to perform local optimization based on their individual data, and then averages all updated model parameters from those devices:
\begin{align}
\label{eq:sim_avg}
	\w^{t+1} = \frac{1}{K}\sum_{k \in S_t} \w^{t+1}_k.
\end{align}

\fedprox{} \cite{tian2019} is a recent improvement of \fedave{} and differs from \fedave{} by adding a proximal term, i.e., $\mu \norm{\w -\w^t}^2$ at round $t$ with proximal parameter $\mu$, to the local objective function, $F_k(\w)$, to regulate the changes made at the local device. Specifically, this proximal term will prevent the updated parameters from deviating too much from the global ones and, hence, from those of other devices as well. \fedprox{} was shown in \cite{tian2019} to have linear convergence rate under assumptions on the near-convex loss functions, dissimilarity between local datasets. In practice, \fedprox{} also performs better than \fedave{} in terms of accuracy, however, the performance is still unstable. Similar idea has also been investigated in a recent work \cite{dinh2020personalized}.

\folb{} \cite{nguyen2020fast} studies the aggregation algorithm at the central server and proposes a smart weighted averaging that takes into account the inner products between local and global gradients with respect to the current model parameters. The underlying motivation is that devices with local gradients similar to the global one give updates in the right directions. On the other hand, for devices with dissimilar local gradients, we should consider the opposite update directions.

We define the context of an optimization round as follows:
\begin{definition}[Context]
	The context of an optimization round $t$ is defined to be the set $\{\w^{t+1}_k, \forall k \in S_t\}$.
\end{definition}
The context captures well the heterogeneity of participating devices in a specific round. Devices with more challenging data (incur large loss by the current model) or those with more computational resources will make major updates, thus, we should consider these more important than the others. \fedave{} and \fedprox{} deploy simple averaging, leading to both slow convergence and unstable performance, while \folb{} considers context with multiple assumptions on near convexity, data dissimilarity, and smoothness, limiting its effectiveness in the majority of practical applications.

\section{Contextual Model Aggregation}
\label{sec:alg}

We consider the same selection of a small set of $K$ devices uniformly at random similarly to \fedave{} and \fedprox{}, however, instead of simply averaging updated parameters from local devices, we propose a contextual aggregation scheme that results in an optimal context-dependent bound.
In the following, we give details of the assumption made (Subsection~\ref{subsec:assumption}), derive context-dependent bound and its optimization (Subsection~\ref{subsec:context_aggregate}), and then extend to the expected bound over random device selection (Subsection~\ref{subsec:expect_aggregate}).

\subsection{Assumptions}
\label{subsec:assumption}
Our proposed method is general, i.e., work for both convex and non-convex loss functions, and make the minimal smoothness assumption that the overall loss function $f(\w)$ is $\beta$-smooth with respect to $\w$. Note that this smoothness assumption is prevalent and essential in the literature of both convex and non-convex optimizations. The following property of smooth functions is extensively used in this work:
\begin{property}[Smoothness]
	\label{prop:smooth}
	Given a $\beta$-smooth function $f(x)$ with respect to $x$ and $x_1, x_2$ in the domain of $f(.)$, we have the following inequality,
	\begin{align}
		f(x_2) \leq f(x_1) + \langle \nabla f(x_1), x_2 - x_1 \rangle + \frac{\beta}{2} \norm{x_2 - x_1}^2.
	\end{align}
\end{property}


\subsection{Deriving an aggregation scheme with optimal context-dependent bound}
\label{subsec:context_aggregate}
We will start from a general form of model aggregation associated with a set of variables and the $\beta$-smoothness property of function $f(\w)$ to obtain a general lower-bound of loss reduction from the context of selected devices $S_t$ in round $t$. Then, we optimize the resulting bound over the aggregation variables to find the optimal setting. We explore practical considerations of parameters involved in the process and show a definite reduction off the loss function after every round of optimization.

Given a set of devices $S_t$ at round $t$ with the corresponding updated parameters $\w^{t+1}_k, \forall k \in S_t$, we first consider a general form of aggregation that adjusts each update $\Delta \w^{t+1}_{k} = \w^{t+1}_k - \w^t$ from device $k$ by a weight parameter $\alpha^t_k$,
\begin{align}
	\w^{t+1} = \w^t + \sum_{k \in S_t} \alpha^t_k \Delta\w^{t+1}_k.
	\label{eq:aggregation}
\end{align}
The weight parameter $\alpha^t_k$ reflects the importance of update from device $k$ in reducing the overall loss function $f(.)$ and must optimally factors in all of the different sources of heterogeneity from data distribution to device's available resources, which translates into how much the device optimizes the parameters based on its local data.

From there, we use the property of $\beta$-smoothness of the function $f(\w)$ to derive the following lower-bound on the loss reduction after round $t$:
\begin{align}
	f(\w^{t+1}) & \leq f(\w^{t}) + \langle \nabla f(\w^{t}), \w^{t+1} - \w^{t}\rangle \nonumber \\
	& \qquad\qquad\qquad+ \frac{\beta}{2} \norm{\w^{t+1} - \w^{t}}^2 \nonumber \\
	& \leq f(\w^{t}) + \langle \nabla f(\w^{t}), \sum_{k \in S_t} \alpha^t_k \Delta \w^{t+1}_k \rangle \nonumber \\
	& \qquad\qquad\qquad+ \frac{\beta}{2} \norm{\sum_{k \in S_t}\alpha^t_k \Delta \w^{t+1}_k}^2.
	\label{eq:loss_bound}
\end{align}
Thus,
\begin{align}
	f(\w^{t}) - f(\w^{t+1}) & \geq -\Bigg( \langle \nabla f(\w^{t}), \sum_{k \in S_t} \alpha_k \Delta \w^{t+1}_k \rangle \nonumber \\
	& \qquad\qquad + \frac{\beta}{2} \norm{\sum_{k \in S_t}\alpha^t_k \Delta \w^{t+1}_k}^2 \Bigg).
	\label{eq:loss_red}
\end{align}
The left hand side of \eqref{eq:loss_red} depicts the reduction of $f(\w)$ after aggregation and the right hand side provides a lower-bound on that reduction. We would want to maximize this lower-bound or, equivalently, minimize the following function 
\begin{align}
	g_t(\pmb{\alpha}^t) = \langle \nabla f(\w^{t}), \sum_{k \in S_t} \alpha^t_k \Delta \w^{t+1}_k \rangle + \frac{\beta}{2} \norm{\sum_{k \in S_t}\alpha^t_k \Delta \w^{t+1}_k}^2, \nonumber
\end{align} 
where $\pmb{\alpha}^t = [\alpha^t_1, \dots, \alpha^t_N]$. This bounding function depends on the updates $\Delta\w^{t+1}_k$ from the current set of devices $S_t$ and weights $\alpha^t_k$. Thus, we call it \emph{context-dependent} and refer to this as the lower-bound function of $\pmb{\alpha}^t$. We will find $\pmb{\alpha}^t$ to make this function as small (negative) as possible.

To derive the optimal $\pmb{\alpha}^t$ that minimizes the lower-bound function, we first look at the partial derivatives of $g_t(\pmb{\alpha}^t)$ with respect to $\alpha^t_{k}$ as follows:
\begin{align}
	\frac{\partial g_t(\pmb{\alpha}^t)}{\partial \alpha^t_k} &= \langle \nabla f(\w^{t}), \Delta \w^{t+1}_k \rangle + \beta \sum_{k' \in S_t}\alpha^t_{k'} \langle \Delta \w^{t+1}_{k}, \Delta \w^{t+1}_{k'}\rangle \nonumber \\
	& = \langle \Delta \w^{t+1}_k , \nabla f(\w^t) + \beta \sum_{k' \in S_t} \alpha^t_{k'}\Delta \w^{t+1}_{k'}\rangle.
\end{align}

The optimal solution satisfies $\frac{\partial g_t(\pmb{\alpha}^t)}{\partial \alpha^t_k}  = 0, \forall k \in S_t$. Thus, $\nabla f(\w^t) + \beta \sum_{k' \in S_t}\alpha^t_{k'}\Delta \w^{t+1}_{k'}$ lives in the \emph{nullspace} of a matrix $\matr{G}_t$ that has $K$ rows corresponding to $\Delta \w^{t+1}_k, \forall k \in S_t$. With presence of various heterogeneity sources, this matrix $\matr{G}_t$ likely has full rank and, hence, its nullspace has rank of $n - K$, where $n$ is the dimension of model parameters $\w$, based on the Rank Plus Nullity Theorem \cite{vandebril2006note} with $K \ll n$. Hence, we need to find $\alpha^t_{k'}, k' \in S_t$ such that  $\nabla f(\w^t) + \beta \sum_{k' \in S_t}\alpha^t_{k'}\Delta \w^{t+1}_{k'}$ lives in the nullspace of $\matr{G}_t$. To achieve that, we first obtain a basis of the nullspace of $\matr{G}_t$ through standard techniques in linear algebra \cite{anton2013elementary}, e.g., SVD. Let the basis consist of vectors $\e_1, \dots, \e_{n-K}$ in $n$ dimension. Since every element in this nullspace is of the form $\sum_{i = 1}^{n-K} x_i \e_i$, we will solve the following equation:
\begin{align}
	\nabla f(\w^t) + \beta \sum_{k' \in S_t}\alpha^t_{k'}\Delta \w^{t+1}_{k'} = \sum_{i = 1}^{n-K} x_i \e_i, 
	\label{eq:alpha_x}
\end{align}
for $\alpha^t_{k'}, k' \in S_t$ and $x_i, i = 1..(n-K)$. Note that \eqref{eq:alpha_x} is a system of $n$ linear equations and $n$ variables and can be solved quickly. The resulting $\alpha^t_{k'}, k' \in S_t$ will be used in our aggregation scheme to obtain a tight (optimal) lower-bound and leads to a better convergence rate.
\begin{proposition}[Optimality]
	The aggregation in \eqref{eq:aggregation} with $\alpha^t_{k'}, k' \in S_t$ being the solution of the system of linear equations in \eqref{eq:alpha_x} produces the optimal context-dependent bound in \eqref{eq:loss_bound}.
\end{proposition}

\textbf{Setting up parameters:}
In the equation \eqref{eq:alpha_x}, we only have $\Delta \w^{t+1}_{k'}$ available and need to determine the settings of smoothness constant $\beta$ and global gradient $\nabla f(\w^t)$.
Since $\beta$ is generally unknown in advance, however, in smooth function optimizations \cite{boyd2004convex}, the ideal learning rate $\l$ is set to $\frac{1}{\beta}$ for fast convergence rate. Thus, we set $\beta = \frac{1}{\l}$ in the above equation \eqref{eq:alpha_x}, where $\l$ is the learning rate of the local optimizers at devices. Moreover, $\nabla f(\w^{t})$ is also generally not available (requiring all the devices to compute), in practice, we select a second set of $K_2$ random devices at each round just to compute the local gradients with respect to $\w^{t}$ and average those local gradients to obtain an estimate of $\nabla f(\w^{t})$. Specifically, in our experiments, we consider and compare various settings of $K_2$ and find that even in the extreme case of $K_2 = 0$, we can just use devices in $S_t$ to obtain an estimate of $\nabla f(\w^t)$ and achieve nearly the same performance as when we use all $N$ devices to compute $\nabla f(\w^t)$. Note also that this estimation step can be done in parallel with the local updates and is much faster than the later as well.

\textbf{Note on efficiency}: For machine learning models with a large number of parameters, e.g., deep neural networks, finding nullspace of $\mathbf{Q}_t$ and solving \eqref{eq:alpha_x} may render some computational challenges at the central server. Fortunately, oftentimes, we can make use of special features of the individual model type to improve efficiency. For example, in deep neural networks, prior studies \cite{ioffe2015batch,ba2016layer,katharopoulos2018not} show that, with standard techniques of weight initialization and activation normalization, variation of gradient/update is mostly captured by that of the last layer's parameters. Hence, we can use a small fraction, corresponding to the last layer's parameters, of the gradient $\nabla f(\w^t)$ and updates $\Delta \w^{t+1}_k$ as good approximations of the full versions.

\begin{algorithm}[h]
	\caption{Contextual aggregation scheme towards optimal convergence.}
	\label{alg:new_aggregation}
	\DontPrintSemicolon
	
	\setstretch{1.2}
	\SetKwInOut{Input}{Input}\SetKwInOut{Output}{output}
	\Input{$\nabla f(\w^t), \Delta \w^{t+1}_{k}, k \in S_t$}
	Compute a basis of the nullspace of $\matr{G}_t$ formed by stacking $\Delta \w^{t+1}_k, \forall k \in S_t$ for rows of this matrix. \\
	Solve the system of linear equations in \eqref{eq:alpha_x} to find $\alpha^t_{k'}, \forall k' \in S_t$ and $x_i, \forall i = 1..(n-k)$. \\
	Use $\alpha^t_{k'}, \forall k' \in S_t$ to compute the next global model parameter set $\w^{t+1}$ according to \eqref{eq:aggregation}. \\
\end{algorithm}

\textbf{Contextual aggregation scheme:} Our new aggregation scheme is described in Algorithm~\ref{alg:new_aggregation}. The first step is to compute a basis of the nullspace of matrix $\matr{G}_t$. Note that this is quick since $\matr{G}_t$ has only a small number of rows, $K$. Then, we solve \eqref{eq:alpha_x} to find $\pmb{\alpha}^t$ that minimizes the lower-bound function, which is used in Line~3 to aggregate $\Delta \w^{t+1}_k$ and form the new set of model parameters $\w^{t+1}$.

With this aggregation scheme in Algorithm~\ref{alg:new_aggregation}, we obtain the following results related to convergence rate of the associated federated learning algorithm.
\begin{theorem}[Definite loss reduction]
	With the assumption that the function $f(\w)$ is $\beta$-smooth, the federated learning algorithm with aggregation scheme in Algorithm~\ref{alg:new_aggregation} provides the following loss reduction at each optimization round:
	\begin{align}
		f(\w^{t}) - f(\w^{t+1}) \geq \frac{\beta}{2} \norm{\sum_{k \in S_t} \alpha^t_{k} \Delta\w_k^{t+1}}^2.
	\end{align}
	\label{thm:converge_loss}
\end{theorem}
\begin{proof}
	Recall that $\alpha^t_{k}, \forall k \in S_t$ satisfy,
	\begin{align}
		\frac{\partial f(\pmb{\alpha}^t)}{\alpha^t_{k}} = \langle \Delta \w^{t+1}_k , \nabla f(\w^t) + \beta \sum_{k' \in S_t} \alpha^t_{k'}\Delta \w^{t+1}_{k'}\rangle = 0.
		\label{eq:solution_identity}
	\end{align}
	Combining the above identity with the lower-bound in \eqref{eq:loss_bound} gives us the following derivation,
	\begin{align}
		& f(\w^{t+1}) \leq f(\w^t) + \langle \nabla f(\w^t), \sum_{k \in S_t} \alpha^t_{k} \Delta\w_k^{t+1} \rangle \nonumber \\
		&\qquad\qquad\qquad + \frac{\beta}{2} \norm{\sum_{k \in S_t} \alpha^t_{k} \Delta\w_k^{t+1}}^2 \nonumber \\
		& = f(\w^t) + \langle \nabla f(\w^t), \sum_{k \in S_t} \alpha^t_{k} \Delta\w_k^{t+1} \rangle \nonumber \\
		&\qquad\qquad\qquad+ \beta \langle \sum_{k\in S_t}\alpha^t_{k} \Delta\w_k^{t+1},\sum_{k' \in S_t} \alpha^t_{k'}\Delta\w_{k'}^{t+1}\rangle \nonumber \\
		&\qquad\qquad\qquad - \frac{\beta}{2} \norm{\sum_{k \in S_t} \alpha^t_{k} \Delta\w_k^{t+1} }^2 \nonumber \\
		&= f(\w^{t}) + \underbrace{\sum_{k \in S_t} \alpha^t_{k} \langle \Delta\w_k^{t+1}, \nabla f(\w^t) + \beta \sum_{k' \in S_t} \alpha^t_{k'}\Delta\w_{k'}^{t+1} \rangle}_{=0 \text{ (due to \eqref{eq:solution_identity})}}  \nonumber \\
		&\qquad\qquad\qquad - \frac{\beta}{2} \norm{\sum_{k \in S_t} \alpha^t_{k} \Delta\w_k^{t+1}t }^2 \nonumber \\
		& = f(\w^{t}) - \frac{\beta}{2} \norm{\sum_{k \in S_t} \alpha^t_{k} \Delta\w_k^{t+1}}^2.
	\end{align}
	That completes the proof.
\end{proof}
\textbf{Remark}: Note that we make no further assumption other than the minimal $\beta$-smoothness of the loss function $f(\w)$. In comparison, previous works commonly rely on convexity, bounded gradient, gradient and Hessian dissimilarity in order to bound loss reduction. Thus, our method is quite general and applicable in a much larger number of cases. Furthermore, the contextual aggregation can be used as a modular plug-and-run component in many existing federated learning algorithms by simply replacing the vanilla aggregation with the contextual one. This allows for great flexibility at the local devices to choose the most suitable optimizer with respect to availability and computing resources. In fact, we demonstrate the combinations of our contextual aggregation with different local optimizations and show significant improved convergence rates compared to that of the existing federated learning algorithms.

The loss reduction in Theorem~\ref{thm:converge_loss} depends primarily on the updates $\Delta \w_k^{t+1}, \forall k \in S_t$, which are the results of local optimization. Hence, if we consider a particular local optimizer, we can bound the updates and further derive a more detailed bound. However, the calculation of $\alpha^t_k$ depends on $\Delta\w_k^{t+1}, \forall k \in S_t$ in a non-trivial way and a clean version of this bound without $\alpha^t_k$ is difficult to obtain. Moreover, with additional assumptions on gradient dissimilarity and SGD as the local optimizer, we can derive formal convergence rate. 

\subsection{Deriving a scheme with optimal expected bound}
\label{subsec:expect_aggregate}
Here we derive an aggregation scheme to maximize the expected loss reduction in round $t$ over the randomness of device selection. The context-dependent bound may vary between different selections of participants and we are interested in optimizing the expected bound over all different choices.  From \eqref{eq:loss_bound}, we take the expectation with respect to the random selection of devices in round $t$ and obtain,
\begin{align}
	\E [f(&\w^{t+1}) ] \leq f(\w^{t}) + \E \left[ \langle \nabla f(\w^{t}), \sum_{k \in S_t} \alpha^t_k \Delta \w^{t+1}_k \rangle \right] \nonumber \\
	& \qquad+ \frac{\beta}{2} \E \left[ \norm{\sum_{k \in S_t}\alpha^t_k \Delta \w^{t+1}_k}^2 \right] \nonumber \\
	& = f(\w^t) + \frac{K}{N} \sum_{k = 1}^{N} \alpha^t_{k} \langle \nabla f(\w^t), \Delta \w^{t+1}_k \rangle \nonumber \\
	& \qquad + \frac{\beta}{2} \E \left[ \sum_{k,k' \in S_t} \alpha^t_k \alpha^t_{k'} \langle \Delta\w_k^{t+1}, \Delta\w_{k'}^{t+1} \rangle \right],
	\label{eq:loss_bound_e}
\end{align}
then,
\begin{align}
	\E[f(\w^t) & - f(\w^{t+1})] \ge - \Bigg( \frac{K}{N}\sum_{k = 1}^{N} \alpha^t_{k} \langle \nabla f(\w^t), \Delta \w^{t+1}_k \rangle\nonumber \\
	&  + \frac{\beta}{2} \E \left[ \sum_{k,k' \in S_t} \alpha^t_k \alpha^t_{k'} \langle \Delta\w_k^{t+1}, \Delta\w_{k'}^{t+1} \rangle \right] \Bigg).
\end{align}

Similarly to the case of context-dependent bound, let $g_t(\pmb{\alpha}^t)$ be the lower-bound function - negated right-hand side of the above inequality, and taking derivatives of $g_t(\pmb{\alpha}^t)$ with respect to $\alpha^t_k$ gives us:
\begin{align}
	&\frac{\partial g(\pmb{\alpha}^t)}{\partial \alpha^t_k} = \frac{K}{N} \langle \nabla f(\w^t), \Delta\w^{t+1}_k \rangle \nonumber \\
	&\qquad\qquad+ \beta \frac{K(K-1)}{N(N-1)} \sum_{k' = 1}^{N} \alpha^t_{k'} \langle \Delta\w_k^{t+1}, \Delta\w_{k'}^{t+1} \rangle \nonumber \\
	&\quad = \langle \Delta\w_k^{t+1}, \frac{K}{N} \nabla f(\w^t) + \beta\frac{K(K-1)}{N(N-1)} \sum_{k' = 1}^{N} \alpha^t_{k'} \Delta\w_{k'}^{t+1} \rangle. \nonumber
\end{align}
We need to solve $\frac{\partial g(\pmb{\alpha}^t)}{\alpha^t_{k}} = 0, \forall k \in 1..N$, and can follow the same footsteps as with the context-dependent bound. However, we would need $\Delta\w_k^{t+1}, \forall k = 1..N$, or, in other words, the full participation of all the devices in the entire network. There are many prior research works \cite{wang19,wang2021device} (see surveys \cite{yang2019federated,li19survey,kairouz19}) that consider the full participation setting and we can apply our techniques to find an optimal $\pmb{\alpha}^t$ in those cases. On the other hand, if this setting of full participation is not feasible, e.g., in the original federated learning framework \cite{mcmahan17}, an approximation can be obtained by sampling a much smaller pool of $N'$ devices in each round of optimization and solve for $\pmb{\alpha}'^t$ that is composed of $\alpha'^t_{k}, \forall k \in S_t$ and $N'$ auxiliary variables $x_i$. The resulting $\alpha'^t_k, \forall k \in S_t$ will then be used in the aggregation scheme. In addition, the settings of $\beta = \frac{1}{\l}$ and an estimate of $\nabla f(\w^t)$ can be obtained in a similar manner as in the context-dependent bound.

Clearly, the resulting bound will not be a definite reduction of loss, but rather, a positive bound on improvement of expected loss. Furthermore, compared to the case of context-dependent bound, solving for $\alpha^t_k$ would require more computation since the matrix $\mathbf{Q}_t$ has $N$ rows, instead of $K$, and finding a basis of $\mathbf{Q}_t$ nullspace is more expensive.

%

\section{Experiments}
\label{sec:exps}
In this section, we present our experimental results to demonstrate the advantages of our contextual aggregation scheme when combined with different existing federated learning algorithms. We show that these combinations offer both faster convergence speed and more robust performance throughout the entire optimization process compared with existing state-of-the-art approaches.

\begin{figure*}[t!]
	\includegraphics[width=\linewidth]{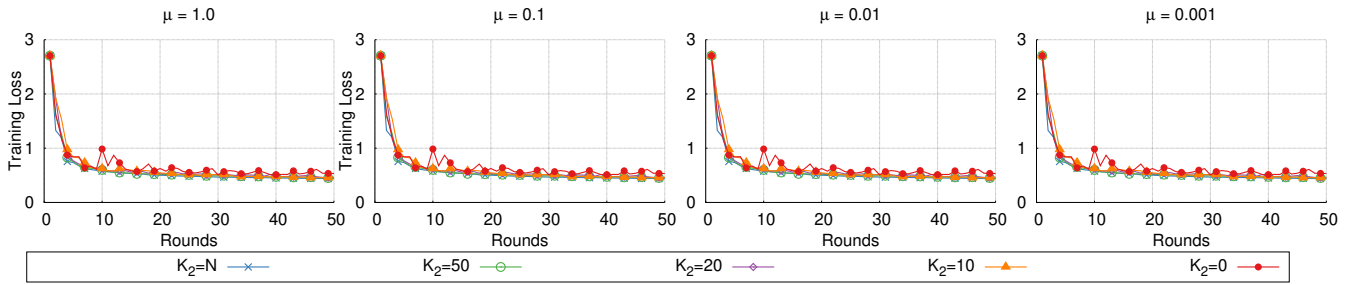}
	\caption{Training loss comparison of different aggregation variants in various settings with \fedproxc{}. All the variants of contextual aggregation deliver similar performance, even in the most extreme case with $K_2 = 0$.}
	\label{fig:variant_compare_loss}
	\vspace{-0.15in}
\end{figure*}
\begin{figure*}[t!]
	\includegraphics[width=\linewidth]{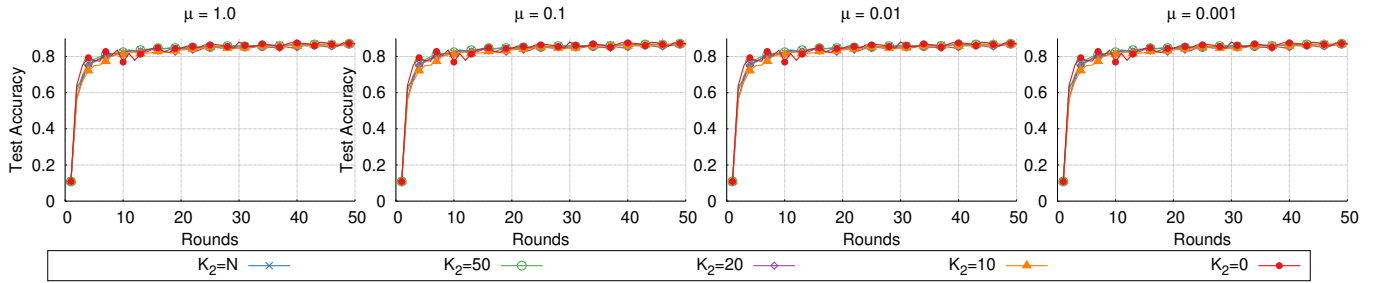}
	\caption{Test accuracy comparison of different aggregation variants in various settings.}
	\label{fig:variant_compare_acc}
	\vspace{-0.15in}
\end{figure*}
\begin{figure*}[t!]
	\includegraphics[width=\linewidth]{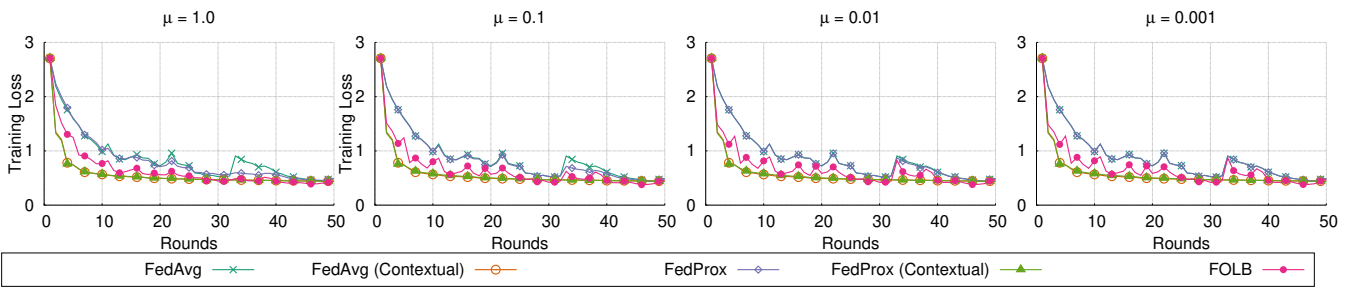}
	\caption{Training loss comparison of different federated learning algorithms in various settings. The contextual versions significantly outperform the existing algorithms in both achieving better performance metrics and robustness between rounds of optimization.}
	\label{fig:compare_loss}
	\vspace{-0.15in}
\end{figure*}
\begin{figure*}[t!]
	\includegraphics[width=\linewidth]{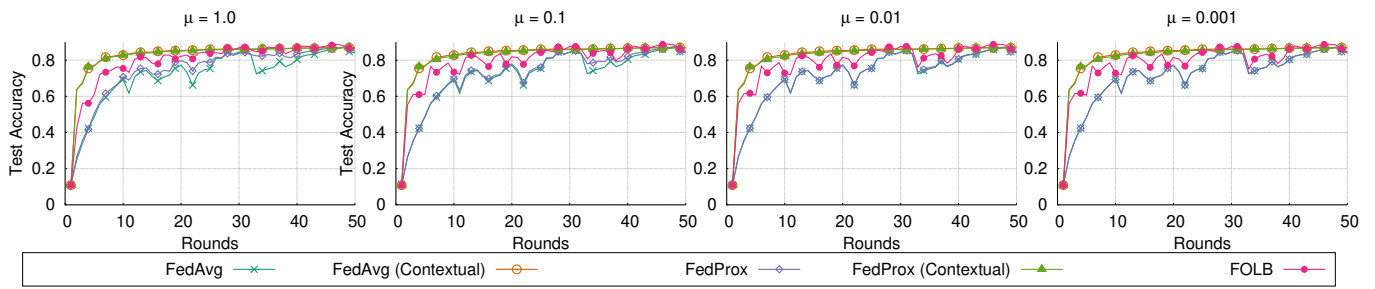}
	\caption{Test accuracy comparison of different federated learning algorithms in various settings.}
	\label{fig:compare_acc}
	\vspace{-0.2in}
\end{figure*}
\begin{figure*}[t!]
	\centering
	\begin{subfigure}[b]{0.87\linewidth}
		\includegraphics[width=\linewidth]{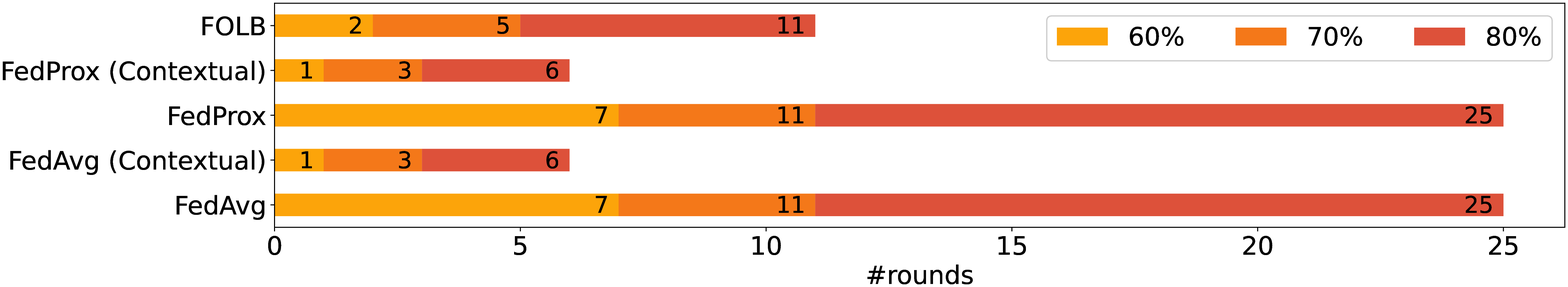}
		\vspace{-0.25in}
		\caption{\mnist}
	\end{subfigure}
	\begin{subfigure}[b]{0.87\linewidth}
		\includegraphics[width=\linewidth]{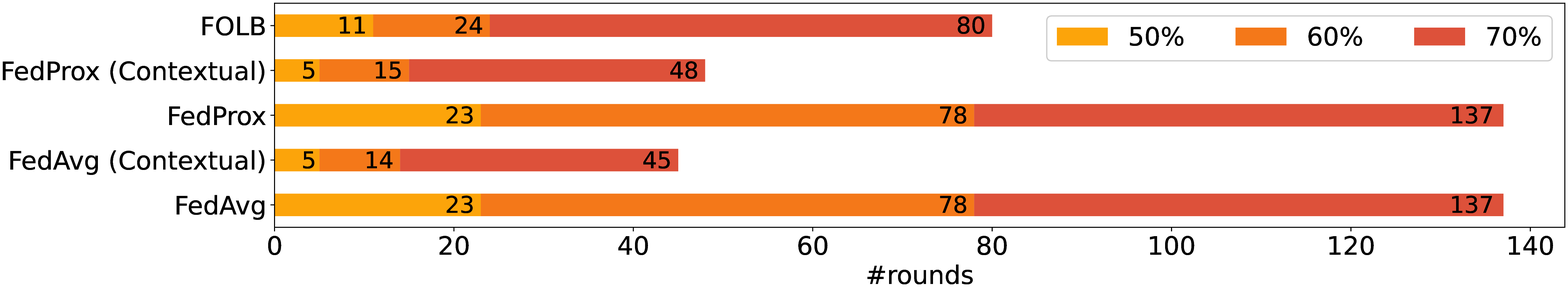}
		\vspace{-0.25in}
		\caption{\emnist}
	\end{subfigure}\\
	\begin{subfigure}[b]{0.87\linewidth}
		\includegraphics[width=\linewidth]{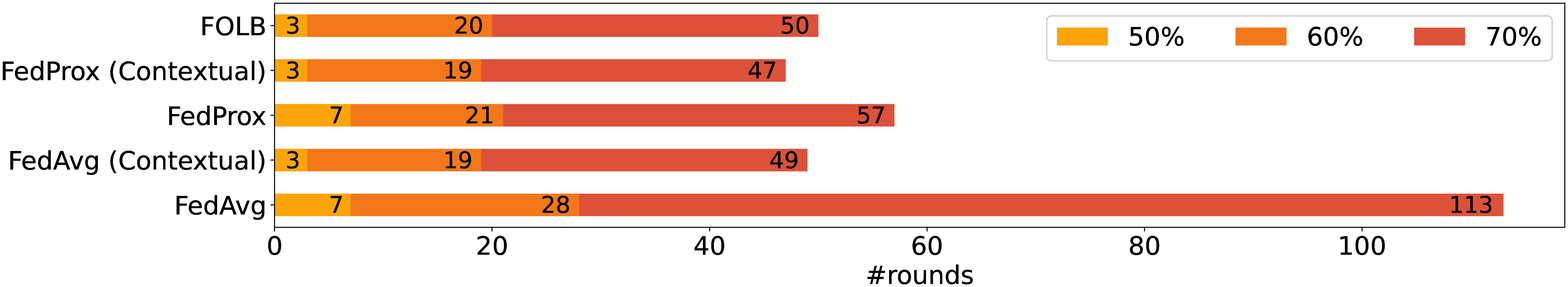}
		\vspace{-0.25in}
		\caption{\syniid}
	\end{subfigure}
	\begin{subfigure}[b]{0.87\linewidth}
		\includegraphics[width=\linewidth]{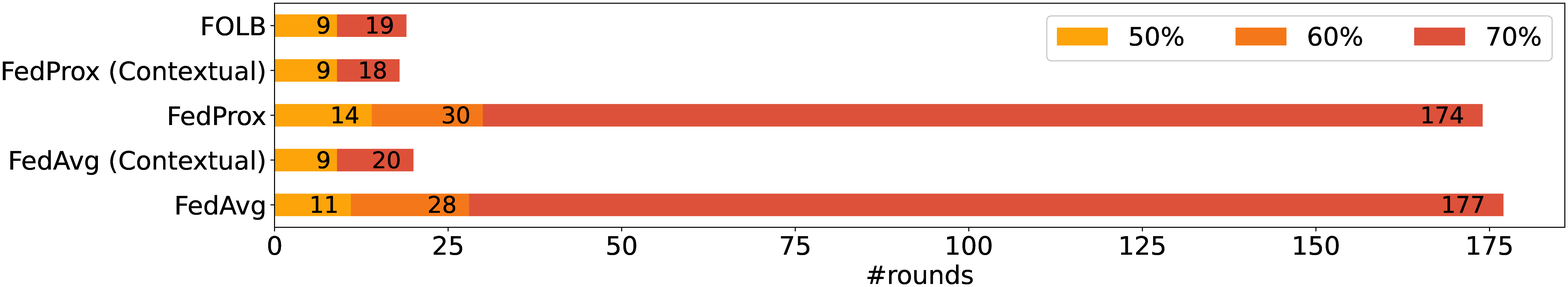}
		\vspace{-0.25in}
		\caption{\synone}
		\vspace{-0.05in}
	\end{subfigure}
	\caption{Number of optimization rounds to reach different accuracy levels required by federated learning algorithms on various datasets ($\mu = 0.1$ for \fedprox{} and variants). The contextual versions require much fewer rounds to achieve the same levels of accuracy compared with the original algorithms.}
	\label{fig:mnist_acc_level}
	\vspace{-0.3in}
\end{figure*}
\subsection{Experimental setup}

\subsubsection{Datasets with data heterogeneity} We adopt four datasets commonly used in the literature \cite{shamir14,li19,nguyen2020fast,li19survey,li19fair}: 10-class \mnist{} \cite{lecun1998gradient}, 62-class Federated Extended \mnist{} (\emnist{}) \cite{cohen2017emnist}, and two synthetic datasets to simulate different levels of data heterogeneity \cite{shamir14,li19}.  The two synthetic datasets, namely \syniid{} and \synone{}, corresponds to scenarios with the i.i.d. data and a high level of data heterogeneity across devices in the network (see \cite{shamir14} for details). We also consider the commonly used multinomial logistic regression as the underlying machine learning model to perform classification tasks on the datasets.

\subsubsection{Federated learning algorithms} To demonstrate the flexibility and effectiveness of the proposed contextual aggregation scheme, we combine it with standard federated learning algorithms by replacing the existing aggregation by the contextual one. In particular, we consider on the canonical \fedave{} \cite{mcmahan17} and denote its contextual version by \fedavec{}, and the recent state-of-the-art \fedprox{}\cite{tian2019} with the contextual variant \fedproxc{}. In addition, we also compare with the recent algorithm \folb{}\cite{nguyen2020fast}, which is equipped with a smart weighted averaging aggregation on top of  \fedprox{}. Thus, \folb{} represents the state-of-the-art aggregation to compare with the proposed scheme. In all the algorithms, we fix the number of devices in each round to $10$, $K = 10$, which is also standard in the literature.

In all the experiments, we consider the more efficient aggregation with context-dependent bound in Subsection~\ref{subsec:context_aggregate}. Moreover, we also examine different variants of the contextual aggregation by varying $K_2$ - the number of devices for estimating $\nabla f(\w^t)$.  Explicitly, we vary $K_2$ between $N$ and $0$, (i.e., $N, 50, 20, 10, 0$). For the case $K_2 = 0$, we just use the same devices in $S_t$ to estimate $\nabla f(\w^t)$.

\subsubsection{Local optimization and computational heterogeneity} We use mini-batch SGD as local optimizer and simulate devices' heterogeneity of computing resources by allowing each device to perform a random number of epochs uniformly selected between 1 and 20. Moreover, all these random selections are kept consistent across all the algorithms and settings by initializing the random number generator for device selection with the same seed.

\subsubsection{Metrics} To study the performance, we report the training loss and test accuracy of the trained model after a certain number of rounds. We also inspect the number of rounds each algorithm requires to achieve a certain level of test accuracy to compare the convergence speeds.

\subsection{Experimental results}
We first study different variants of our contextual aggregation scheme and then compare them with the existing \fl{} algorithms in terms of performance and robustness.
\subsubsection{Comparison between contextual variants}
This first set of experiments focuses on variants of the proposed contextual aggregation scheme with varying $K_2$ - number of devices to estimate the global gradient $\nabla f(\w^t)$ in each round $t$. We show that a very small number of devices is sufficient to achieve the top performance of the contextual aggregation. We use \fedproxc{} with various settings of the proximal parameter $\mu$ (see \cite{tian2019} for more details), since proximal local optimization has witnessed numerous successes in dealing with data heterogeneity in federated learning \cite{tian2019,dinh2020personalized}. These experiments are run on \mnist{} dataset.

Figures~\ref{fig:variant_compare_loss} and ~\ref{fig:variant_compare_acc} display the training loss and test accuracy of  the examined variants. The results reveal an interesting characteristic that even with a very small value of $K_2$, the performance is still near the peak at $K_2 = N$. Specifically, the variants $K_2 = N$, $K_2 = 50$, $K_2 = 20$, and $K_2 = 10$, have visually indistinguishable performance in both training loss and test accuracy. We only observe some differences in the extreme case of $K_2 = 0$, where we need to use the same set of devices $S_t$ to both estimate $\nabla f(\w^t)$ and perform local updates. However, the differences are negligible and appear as minor fluctuations from the ideal case of $K_2 = N$. The reason for these differences is likely due to correlation between the estimated $\nabla f(\w^t)$ and the updated parameters from the same set of devices. This correlation is broken when $K_2 > 0$ and our aggregation performs near optimality as illustrated by the results of other variants.

Another observation is the fast convergence and robust performance of all the variants. They all converge after about 20 rounds of optimization and remain at the same peak level of minimum loss and maximum accuracy. There is also not much difference between the results with distinct values of proximal parameter $\mu$ indicating that any advantage in performance will be attributed to the core contextual aggregation and not to the proximal terms added to local optimization. Henceforth, we would expect \fedavec{} and \fedproxc{} to  perform similarly in subsequent experiments.

\begin{figure}[h!]
	\centering
	\includegraphics[width=0.99\linewidth]{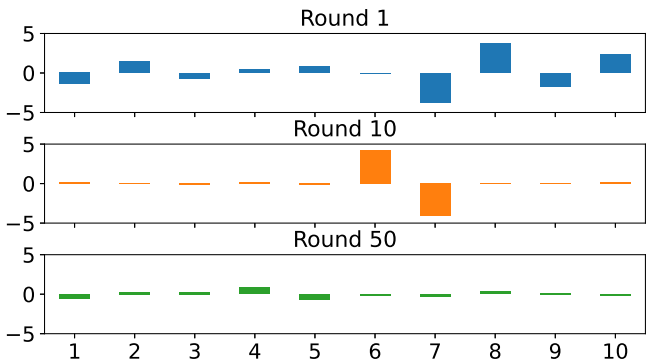}
	\caption{Vastly different aggregation variables computed at the early, near-converged and converged stages.}
	\label{fig:learn_weights}
	\vspace{-0.25in}
\end{figure}
\subsubsection{Optimal aggregation variables}
In Figure~\ref{fig:learn_weights}, we plots the computed aggregation variables $\alpha^t_k, k \in S_t$ at different rounds corresponding to three stages of the optimization process, i.e., very early, near-converged, and converged. It is clear that aggregation variables vary substantially between values and stages. In early stage, most updates are valuable in different directions. In near-converged stage, only a few updates are necessary, while, at convergence, the updates have roughly the same role. This is in contrast with the common simple averaging in \fedave{}, \fedprox{} and most prior works.

\subsubsection{Comparison between contextual federated learning and existing algorithms}
We compare the contextual versions, \fedavec{} and \fedproxc{},  with the original \fedave{} and \fedprox{}, and the recent \folb{} with a smart aggregation scheme.

Figures~\ref{fig:compare_loss} and~\ref{fig:compare_acc} illustrate training loss and test accuracy of all the considered algorithms on \mnist{} dataset. In both metrics, the contextual versions noticeably converge faster and are robust throughout the optimization rounds than both \fedave{}, \fedprox{}, and \folb{} in all the experiments. The curves produced by contextual aggregation are always better (lower in loss and higher in accuracy) than the rest and very smooth. As also shown in numerous studies \cite{tian2019,nguyen2020fast,li19survey}, \fedave{} and \fedprox{} show great fluctuations in performance even between consecutive rounds due to data heterogeneity. \folb{} achieves better results than \fedave{} and \fedprox{} in individual rounds, however, it suffers from the same fluctuations, especially with smaller values of the proximal constant $\mu$. This shows the ineffectiveness of the aggregation algorithm of \folb{} in dealing with the robustness issue in federated learning. In contrast, our contextual versions provide both higher results and great robustness in improving performance after each round or maintaining the same peak level after converged.
The results here experimentally verify the property of our contextual aggregation that it definitely reduces the loss value for a broad class of functions.

As expected, \fedavec{} and \fedproxc{} perform markedly the same. This shows that contextual aggregation addresses well data heterogeneity and nullifies the effect of proximal term. On the other hand, \fedprox{} with large $\mu$ shows some margins to \fedave{} thanks to the proximal term in local optimization to mitigate data heterogeneity as shown in \cite{tian2019}. However, we also take into account the inherent computing resource heterogeneity across local devices, thus, both \fedave{} and \fedprox{} exhibit slow convergence and ample performance instability between rounds.

Figure~\ref{fig:mnist_acc_level} gives a broader view of the number of  rounds required by the tested federated learning algorithms to attain various levels of accuracy, i.e., 50\%, 60\%, 70\%, and 80\%, on all the datasets to demonstrate the convergence speed. On the two real datasets, \mnist{} and \emnist{}, the contextual versions reduce the number of rounds required by the original algorithms by a factor of 3 or more consistently across all levels of accuracy. Compared to \folb{}, our contextual algorithms also reduce the number of rounds by 2 times. On the i.i.d. dataset \syniid, \folb{}, \fedprox{}, and the two contextual algorithms perform similarly and are much better than \fedave{}. This shows that in the i.i.d. case, most algorithms already achieve good performance. However, in the highly non-i.i.d. case of \synone, both \fedave{} and \fedprox{} fall short again due to the issues with data heterogeneity. \folb{} and our contextual algorithms all require small number of rounds to reach the specified accuracy levels. Hence, both the weighted aggregation in \folb{} and our contextual scheme work well in this synthetic dataset with a high level of heterogeneity.

\section{Related works}
Federated learning falls under the umbrella of distributed optimization, which has been studied extensively in the literature \cite{boyd2011distributed,smith16}. However, early works on distributed optimization focused on datacenter or cloud-based scenarios where shareable data are distributed across a relatively small number of nearby powerful machines, and communication cost as well as privacy are not an issue in traditional setup. However, federated learning targets the massive networks of mobile devices with much slower communication and data privacy is a primary concern. A vast amount of work \cite{mcmahan17,yu18,jiang18,wang19,tian2019,li19survey,dinh2020personalized,karimireddy2020scaffold,kairouz2019advances,khaled2020tighter} has studied the new challenges and optimization algorithms in this setting.

\fedave{}\cite{mcmahan17} is the first and likely most popular federated learning algorithm due to its simplicity and effectiveness. Different from earlier works in distributed optimization, \fedave{} employs multiple epochs of local SGD as optimizer in each round. The updated parameters from a small random set of devices are averaged to serve as the new global model. Later works in \cite{li19,yu19,khaled2020tighter,tian2019,karimireddy2020scaffold} analyzed convergence rate of \fedave{} under assumptions on convexity, bounded gradient, and bounded dissimilarity between devices. Other works \cite{tian2019,dinh2020personalized} improved over \fedave{} by introducing a proximal term into local loss functions to mitigate data heterogeneity and make the loss strongly convex. With similar goal, the work in \cite{karimireddy2020scaffold} introduces a control variate to local updates as a correction mechanism to combat data heterogeneity. The issue with control variate is that all the devices are \emph{stateful} that means each device needs to maintain a control variate the entire time and may not even be used due to random selection of devices to participate. In contrast to these works, we allow any local optimizer to be used under the effect of devices' heterogeneity, and focus our attention on the more resourceful central server, that performs aggregation of local updates. With more computing capability, the server should be able to evaluate the usefulness of each update and combine them in an optimal way that maximally reduces the overall loss function.

Most related to our work is the recent \folb{} algorithm in \cite{nguyen2020fast}. \folb{} builds on top of \fedprox{} \cite{li19} and subsumes a smart aggregation scheme that weighs each local update by the inner product between the local and global gradients. This inner product effectively determines whether the update performed at the device are on the right direction with respect to the global gradient. In cases where 1) the loss function is smooth, 2) local losses are near convex, and 3) gradient dissimilarity between devices is founded, \folb{} can reduce the expected loss and improve over \fedprox{}. Different from \folb{}, the proposed contextual aggregation scheme only uses the smoothness assumption of the overall loss function while providing definite loss reduction in every round of optimization. Our experimental results in Section~\ref{sec:exps} show significantly better convergence speed and robustness of our contextual aggregation scheme over that in \folb{}.

Other related works include \cite{wang19,wang2021device,lin2021two} that consider full participation of all devices in every round and also study offloading data to improve convergence speed, \cite{jeon2020gradient,amiriconvergence,chen2019joint,chen2020convergence} which studies federated learning over wireless networks. Comprehensive surveys can be found in \cite{li19survey,kairouz2019advances}.

\section{Conclusion}
We have proposed contextual aggregation algorithms in federated learning that obtains an optimal bound on loss reduction and demonstrated both faster convergence speed and high level of robustness during the optimization process. The proposed aggregations can be integrated into many existing algorithms and provide significant improvements over the vanilla versions. We have considered multiple variants of the proposed approach and compared them with state-of-the-art federated learning algorithms to illustrate the advantages of our solutions.

Our future works include implementing the contextual federated learning algorithms in practical edge computing systems and investigating performance under this setting as well as possible issues arisen in practical systems.

\bibliographystyle{IEEEtran}
\bibliography{fedlearn}

\begin{thebibliography}{10}
\providecommand{\url}[1]{#1}
\csname url@samestyle\endcsname
\providecommand{\newblock}{\relax}
\providecommand{\bibinfo}[2]{#2}
\providecommand{\BIBentrySTDinterwordspacing}{\spaceskip=0pt\relax}
\providecommand{\BIBentryALTinterwordstretchfactor}{4}
\providecommand{\BIBentryALTinterwordspacing}{\spaceskip=\fontdimen2\font plus
\BIBentryALTinterwordstretchfactor\fontdimen3\font minus
  \fontdimen4\font\relax}
\providecommand{\BIBforeignlanguage}[2]{{%
\expandafter\ifx\csname l@#1\endcsname\relax
\typeout{** WARNING: IEEEtran.bst: No hyphenation pattern has been}%
\typeout{** loaded for the language `#1'. Using the pattern for}%
\typeout{** the default language instead.}%
\else
\language=\csname l@#1\endcsname
\fi
#2}}
\providecommand{\BIBdecl}{\relax}
\BIBdecl

\bibitem{mcmahan17}
B.~McMahan, E.~Moore, D.~Ramage, S.~Hampson, and B.~A. y~Arcas,
  ``Communication-efficient learning of deep networks from decentralized
  data,'' in \emph{Proceedings of the Artificial Intelligence and
  Statistics}.\hskip 1em plus 0.5em minus 0.4em\relax PMLR, 2017, pp.
  1273--1282.

\bibitem{li19survey}
T.~Li, A.~K. Sahu, A.~Talwalkar, and V.~Smith, ``Federated learning:
  Challenges, methods, and future directions,'' 2019, arXiv preprint
  arXiv:1908.07873.

\bibitem{yang2019federated}
Q.~Yang, Y.~Liu, Y.~Cheng, Y.~Kang, T.~Chen, and H.~Yu, ``Federated learning,''
  \emph{Synthesis Lectures on Artificial Intelligence and Machine Learning},
  vol.~13, no.~3, pp. 1--207, 2019.

\bibitem{kairouz2019advances}
P.~Kairouz, H.~B. McMahan, B.~Avent, A.~Bellet, M.~Bennis, A.~N. Bhagoji,
  K.~Bonawitz, Z.~Charles, G.~Cormode, R.~Cummings \emph{et~al.}, ``Advances
  and open problems in federated learning,'' 2019, arXiv preprint
  arXiv:1912.04977.

\bibitem{li2019convergence}
X.~Li, K.~Huang, W.~Yang, S.~Wang, and Z.~Zhang, ``On the convergence of fedavg
  on non-iid data,'' 2019, arXiv preprint arXiv:1907.02189.

\bibitem{yu2019parallel}
H.~Yu, S.~Yang, and S.~Zhu, ``Parallel restarted sgd with faster convergence
  and less communication: Demystifying why model averaging works for deep
  learning,'' in \emph{Proceedings of the AAAI Conference on Artificial
  Intelligence}, vol.~33, no.~01, 2019, pp. 5693--5700.

\bibitem{khaled2020tighter}
A.~Khaled, K.~Mishchenko, and P.~Richt{\'a}rik, ``Tighter theory for local sgd
  on identical and heterogeneous data,'' in \emph{Proceedings of the
  International Conference on Artificial Intelligence and Statistics}.\hskip
  1em plus 0.5em minus 0.4em\relax PMLR, 2020, pp. 4519--4529.

\bibitem{tian2019}
T.~Li, A.~K. Sahu, M.~Zaheer, M.~Sanjabi, A.~Talwalkar, and V.~Smith,
  ``Federated optimization in heterogeneous networks,'' 2020, arXiv preprint
  arXiv:1812.06127.

\bibitem{dinh2020personalized}
C.~T. Dinh, N.~H. Tran, and T.~D. Nguyen, ``Personalized federated learning
  with moreau envelopes,'' 2020, arXiv preprint arXiv:2006.08848.

\bibitem{karimireddy2020scaffold}
S.~P. Karimireddy, S.~Kale, M.~Mohri, S.~Reddi, S.~Stich, and A.~T. Suresh,
  ``Scaffold: Stochastic controlled averaging for federated learning,'' in
  \emph{Proceedings of the International Conference on Machine Learning}.\hskip
  1em plus 0.5em minus 0.4em\relax PMLR, 2020, pp. 5132--5143.

\bibitem{nguyen2020fast}
H.~T. Nguyen, V.~Sehwag, S.~Hosseinalipour, C.~G. Brinton, M.~Chiang, and H.~V.
  Poor, ``Fast-convergent federated learning,'' \emph{IEEE Journal on Selected
  Areas in Communications}, vol.~39, no.~1, pp. 201--218, 2020.

\bibitem{tu20}
Y.~Tu, Y.~Ruan, S.~Wagle, C.~G. Brinton, and C.~Joe-Wong, ``Network-aware
  optimization of distributed learning for fog computing,'' in
  \emph{Proceedings of the IEEE Conference on Computer Communications}.\hskip
  1em plus 0.5em minus 0.4em\relax IEEE, 2020, pp. 2509--2518.

\bibitem{vandebril2006note}
R.~Vandebril and M.~Van~Barel, ``A note on the nullity theorem,'' \emph{Journal
  of computational and applied mathematics}, vol. 189, no. 1-2, pp. 179--190,
  2006.

\bibitem{anton2013elementary}
H.~Anton and C.~Rorres, \emph{Elementary linear algebra: applications
  version}.\hskip 1em plus 0.5em minus 0.4em\relax John Wiley \& Sons, 2013.

\bibitem{boyd2004convex}
S.~Boyd, S.~P. Boyd, and L.~Vandenberghe, \emph{Convex optimization}.\hskip 1em
  plus 0.5em minus 0.4em\relax Cambridge university press, 2004.

\bibitem{ioffe2015batch}
S.~Ioffe and C.~Szegedy, ``Batch normalization: Accelerating deep network
  training by reducing internal covariate shift,'' in \emph{Proceedings of the
  International Conference on Machine Learning}.\hskip 1em plus 0.5em minus
  0.4em\relax PMLR, 2015, pp. 448--456.

\bibitem{ba2016layer}
J.~L. Ba, J.~R. Kiros, and G.~E. Hinton, ``Layer normalization,'' 2016, arXiv
  preprint arXiv:1607.06450.

\bibitem{katharopoulos2018not}
A.~Katharopoulos and F.~Fleuret, ``Not all samples are created equal: Deep
  learning with importance sampling,'' in \emph{Proceedings of the
  International Conference on Machine Learning}.\hskip 1em plus 0.5em minus
  0.4em\relax PMLR, 2018, pp. 2525--2534.

\bibitem{wang19}
S.~Wang, T.~Tuor, T.~Salonidis, K.~K. Leung, C.~Makaya, T.~He, and K.~Chan,
  ``Adaptive federated learning in resource constrained edge computing
  systems,'' \emph{IEEE Journal on Selected Areas in Communications}, vol.~37,
  no.~6, pp. 1205--1221, 2019.

\bibitem{wang2021device}
S.~Wang, M.~Lee, S.~Hosseinalipour, R.~Morabito, M.~Chiang, and C.~G. Brinton,
  ``Device sampling for heterogeneous federated learning: Theory, algorithms,
  and implementation,'' 2021, arXiv preprint arXiv:2101.00787.

\bibitem{kairouz19}
P.~Kairouz, H.~B. McMahan, B.~Avent, A.~Bellet, M.~Bennis, A.~N. Bhagoji,
  K.~Bonawitz, Z.~Charles, G.~Cormode, R.~Cummings \emph{et~al.}, ``Advances
  and open problems in federated learning,'' 2019, arXiv preprint
  arXiv:1912.04977.

\bibitem{shamir14}
O.~Shamir, N.~Srebro, and T.~Zhang, ``Communication-efficient distributed
  optimization using an approximate newton-type method,'' in \emph{Proceedings
  of the International Conference on Machine Learning}.\hskip 1em plus 0.5em
  minus 0.4em\relax PMLR, 2014, pp. 1000--1008.

\bibitem{li19}
X.~Li, K.~Huang, W.~Yang, S.~Wang, and Z.~Zhang, ``On the convergence of fedavg
  on non-iid data,'' 2019, arXiv preprint arXiv:1907.02189.

\bibitem{li19fair}
T.~Li, M.~Sanjabi, and V.~Smith, ``Fair resource allocation in federated
  learning,'' 2019, arXiv preprint arXiv:1905.10497.

\bibitem{lecun1998gradient}
Y.~LeCun, L.~Bottou, Y.~Bengio, and P.~Haffner, ``Gradient-based learning
  applied to document recognition,'' \emph{Proceedings of the IEEE}, vol.~86,
  no.~11, pp. 2278--2324, 1998.

\bibitem{cohen2017emnist}
G.~Cohen, S.~Afshar, J.~Tapson, and A.~Van~Schaik, ``Emnist: Extending mnist to
  handwritten letters,'' in \emph{Proceedings of the 2017 International Joint
  Conference on Neural Networks}.\hskip 1em plus 0.5em minus 0.4em\relax IEEE,
  2017, pp. 2921--2926.

\bibitem{boyd2011distributed}
S.~Boyd, N.~Parikh, and E.~Chu, \emph{Distributed Optimization and Statistical
  Learning via the Alternating Direction Method of Multipliers}.\hskip 1em plus
  0.5em minus 0.4em\relax Now Publishers Inc, 2011.

\bibitem{smith16}
V.~Smith, S.~Forte, C.~Ma, M.~Takac, M.~I. Jordan, and M.~Jaggi, ``Cocoa: A
  general framework for communication-efficient distributed optimization,''
  2016, arXiv preprint arXiv:1611.02189.

\bibitem{yu18}
H.~Yu, S.~Yang, and S.~Zhu, ``Parallel restarted sgd for non-convex
  optimization with faster convergence and less communication,'' 2018, arXiv
  preprint arXiv:1807.06629.

\bibitem{jiang18}
P.~Jiang and G.~Agrawal, ``A linear speedup analysis of distributed deep
  learning with sparse and quantized communication,'' in \emph{Proceedings of
  the 32nd International Conference on Neural Information Processing Systems},
  2018, pp. 2530--2541.

\bibitem{yu19}
H.~Yu, R.~Jin, and S.~Yang, ``On the linear speedup analysis of communication
  efficient momentum sgd for distributed non-convex optimization,'' 2019, arXiv
  preprint arXiv:1905.03817.

\bibitem{lin2021two}
F.~P.-C. Lin, S.~Hosseinalipour, S.~S. Azam, C.~G. Brinton, and N.~Michelusi,
  ``Two timescale hybrid federated learning with cooperative d2d local model
  aggregations,'' 2021, arXiv preprint arXiv:2103.10481.

\bibitem{jeon2020gradient}
Y.-S. Jeon, M.~M. ~, J.~Li, and H.~V. Poor, ``Gradient estimation for federated
  learning over massive mimo communication systems,'' 2020, arXiv preprint
  arXiv:2003.08059.

\bibitem{amiriconvergence}
M.~M. Amiri, D.~G{\"u}nd{\"u}z, S.~R. Kulkarni, and H.~V. Poor, ``Convergence
  of update aware device scheduling for federated learning at the wireless
  edge,'' \emph{IEEE Transactions on Wireless Communications}, vol.~20, no.~6,
  pp. 3643--3658, 2021.

\bibitem{chen2019joint}
M.~Chen, Z.~Yang, W.~Saad, C.~Yin, H.~V. Poor, and S.~Cui, ``A joint learning
  and communications framework for federated learning over wireless networks,''
  2019, arXiv preprint arXiv:1909.07972.

\bibitem{chen2020convergence}
M.~Chen, H.~V. Poor, W.~Saad, and S.~Cui, ``Convergence time optimization for
  federated learning over wireless networks,'' 2020, arXiv preprint
  arXiv:2001.07845.

\end{thebibliography}

\end{document}